# Using Deep Learning to Automate the Detection of Flaws in Nuclear Fuel Channel UT Scans

Issam Hammad, Ryan Simpson, Hippolyte Djonon Tsague, and Sarah Hall

*Abstract*— Nuclear reactor inspections are critical to ensure the safety and reliability of a nuclear facility's operation. In Canada, Ultrasonic Testing (UT) is used to inspect the health of pressure tubes which are part of Canada's Deuterium Uranium (CANDU™) reactor's fuel channels. Currently, analysis of UT scans is performed by manual visualization and measurement to locate, characterize, and disposition flaws. Therefore, there is motivation to develop an automated method that is fast and accurate. In this paper, a proof of concept (PoC) that automates the detection of flaws in nuclear fuel channel UT scans using a convolutional neural network (CNN) is presented. The CNN model was trained after constructing a dataset using historical UT scans and the corresponding inspection results. The requirement for this prototype was to identify the location of at least a portion of each flaw in UT scans while minimizing false positives (FPs). The proposed CNN model achieves this target by automatically identifying at least a portion of each flaw where further manual analysis is performed to identify the width, the length, and the type of the flaw.

*Index Terms*— Convolutional Neural Network (CNN), Deep Learning, Fuel Channel Inspection, Non-destructive examination (NDE), Nuclear Reactor, Pipe Inspection, Process Automation, Ultrasonic Testing (UT)

## I. Introduction

THE inspection and maintenance of CANDU™ (Canada Deuterium Uranium) reactors fuel channels is a critical task during outages. The pressure tubes, in particular, are inspected for flaws to ensure the safety and reliability of these reactors [1-5]. Different flaw types such as debris fret, crevice corrosion, and fuel bundle bearing pad fretting (FBBPF) can occur over time. Pressure tubes are inspected using Ultrasonic Testing (UT), a non-destructive examination (NDE) technique, to assess fitness for service. The usage of NDE in the inspection of nuclear power plants has been previously discussed in [5-10]. CANDU™ reactors contain 380-480 fuel channels. Each fuel channel contains a pressure tube which is surrounded by a calandria tube. During operation, new fuel bundles are loaded into the fuel channel, and old fuel bundles are removed. During an inspection, remotely operated tooling is utilized to deliver inspection heads into defueled fuel channels for scanning. The inspection heads are pushed axially down the fuel channel as they are rotated, resulting in a helical collection of UT data for each inspected fuel channel. Fig. 1 illustrates a high-level design for the reactor core and the fuel channels.

The collected UT data (B-scans) are represented as a three-dimensional graphic of UT data with ultrasound travel time as one axis (X-axis), transducer displacement as the second (y-axis), and signal amplitude as the third (shading). B-scans contain a significant amount of data and can have file sizes in the gigabyte range. Currently, collected scans are analyzed manually by highly trained and certified UT analysts. The current process for fuel channel UT analysis involves a manual visual inspection of the data to identify the location of pressure tube flaws. Additionally, a recent inspection tooling modification has introduced mechanical noise (chatter) into the UT collected data. This has further increased the amount of time required to analyze the scanned data. Fig. 2 illustrates an example of a portion of a flaw-free scan with and without mechanical chatter.

Automating the detection of flaws and their locations has enormous advantages in terms of reducing outage duration and operational costs. Therefore, building a machine learning solution that can automate the detection of flaws in CANDU™ fuel channels is needed. This PoC used a large dataset consisting of previous UT scans and the corresponding inspection results. Each UT scan file can be seen as a highly unbalanced dataset as the vast majority of each scan is made up of flaw-free portions. The requirement for this prototype was to identify the location of at least a portion of each flaw in fuel channel scans while minimizing false positives (FPs). Therefore, avoiding false negatives (FNs) has a higher priority over minimization of FPs.

Deep learning, which is a sub-field of machine learning, can be defined as using a computational model that consists of multiple processing layers to learn representations of data with multiple levels of abstraction. We refer to this model as a deep neural network (DNN) [11]. Previously, deep learning [11-12] was utilized in various NDE inspections, as can be seen in [13-15]. Additionally, deep learning for other types of pipe inspections, such as sewer pipes has been discussed in [16-19]. Convolutional neural networks (CNNs) are one type of DNNs designed to process data that are formed as multiple arrays. CNNs are utilized to solve computer vision problems, with image classification being one of the most popular applications [11], [20-21]. The primary advantage of a CNN over a DNN is the concept of weight sharing, where the same set of filters can

Manuscript received in May 2021, revised in August 2021. This work was supported by the Engineering Department at Alithya Digital Technology Corporation, Pickering, ON, Canada

I. Hammad, R Simpson, H. Tsague. and S. Hall are with the Department of the Energy Department at Alithya Digital Technology, 1420 Bayly St, Pickering, ON L1W 3R3, Canada.

I. Hammad, is also with the Department of Electrical and Computer Engineering, Dalhousie University. Halifax, NS B3J 2X4, Canada. (Issam.hammad@dal.ca)

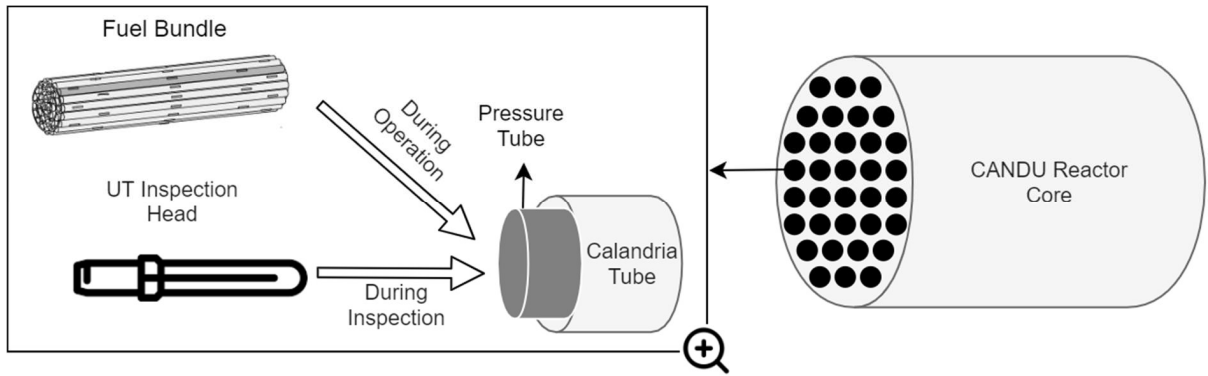

Fig. 1 High-level design for the reactor core and the fuel channels.

be applied to different parts of the image. CNNs usually consist of a large number of filters that are applied to the input image to construct a feature map. In this paper, a CNN model is proposed to automate the detection of flaws in CANDU™ pressure tubes. Using a CNN model allows for an automated construction of a feature map that represent different flaw shapes. Such a feature map can identify flaws even with the existence of the chatter in some UT Scans. The CNN model is followed by a deterministic post-processor which reduces False Positives (FPs) as the paper will present.

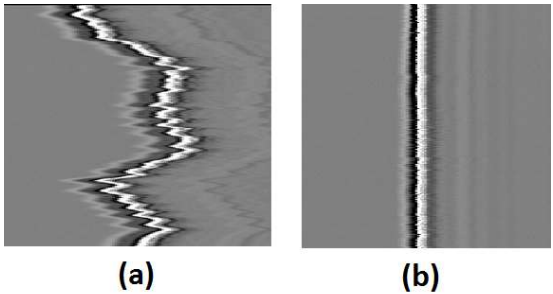

**(a)**      **(b)**

Fig. 2. A portion of UT B-Scan (a) with mechanical chatter (b) without mechanical chatter.

This paper is structured as follows: Section II provides background on the data and the application. In Section III, the steps for creating the training dataset are demonstrated. Section IV presents the proposed CNN model. Section V presents test results for various UT scan files. Section VI summarizes the research conclusion.

## II. BACKGROUND

During the PoC, a large set of historical UT scan files were utilized that include data from different nuclear stations during different outages that occurred in the years 2013-2018. Each scan file represents a stack of ultrasonic waves sent by the probe to a specific axial and rotary position inside the pressure tube. Each ultrasonic wave has a time of flight (TOF) and an amplitude representing the time it met the pressure tube surface. Fig. 3 demonstrates an example of one ultrasonic wave. Most UT scan files are captured with a resolution of 0.2mm/0.1°. A 1mm scan portion will have on average 18,000 captured ultrasonic waves.

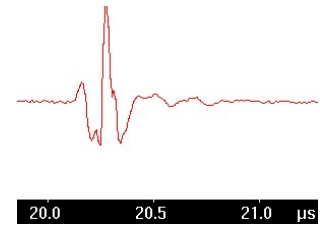

Fig. 3. An ultrasonic scan response for the inspection of a specific axial and rotary position.

UT scan files vary in size depending on the scanned length and the rotary range. Lengths of 20-50mm in each file are typical. Some files contain a full rotary scan from 0-359.9 degrees, while other scan files contain a specific rotary range. The received scan file sizes varied between 50MB and 6GB. Most ultrasonic waves in the UT scan files represent a pressure tube portion that is flaw-free; therefore, scan files can be seen as a highly unbalanced dataset. Determining if a specific region of the pressure tube has a flaw is currently performed by manual analysis by calculating the time difference between the peaks of the suspected flaw region and the peak of a healthy region. If the time difference translates to a calculated depth that is greater than 0.1mm a flaw is recorded in that region. In certain instances, two healthy reference signals are used with linear interpolation. Fig. 4 shows an example for flaw depth measurement.

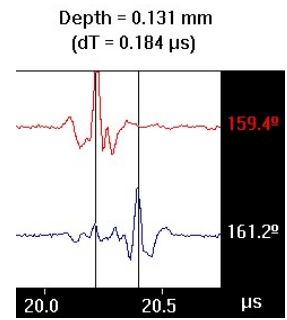

Fig. 4. Flaw depth measurement

Each ultrasonic wave inspects only a portion of a pressure tube with a resolution of 0.2mm/0.1°. Flaw regions in terms of length and width far exceed this resolution. Therefore, in most cases, multiple ultrasonic waves which exceed the defined depth region exist in each flaw region. For example, a debris fret flaw can have a length of 2.5-4mm with a width ranging between 1-4°. Other, much larger flaws can have lengths that are greater than 20mm. In each UT scan file, less than 0.25% of waveforms represent a flaw indication, and in some files, the ratio is less than 0.01%. This highly unbalanced ratio makes it difficult to build a machine learning model that predicts flaws at the resolution of 0.2mm/0.1°. Such a model would require an extremely high specificity to avoid a flood of FPs. To solve this problem, processing at a much lower resolution can be applied. This can be done by treating this problem as a computer vision problem and stacking bundles of 20 waveforms together to represent a resolution of 0.2mm/2°. Fig. 5 shows an example of a grayscale image of 20 stacked waveforms.

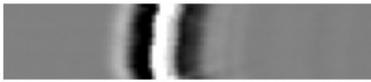

Fig. 5. A stack of 20 waveforms presented as a grayscale image

For the CNN inputs, a 5-channel tensor with 20 waveforms in each channel was used. Therefore, processing could occur at a resolution of 1mm/2° instead of 0.2mm/0.1°. The next chapter discusses in detail the dataset creation steps and provides examples for the constructed inputs.

## III. Creating the Training Dataset

For training and cross-validation, using the UT scan files directly is not feasible as they are highly unbalanced. Therefore, creating a custom balanced training dataset is required. This dataset was created using data from 181 UT scan files and included a total of 60,000 input points representing a resolution of 1mm/2°. The dataset was split into 50,000 training points and 10,000 cross-validation points. The input points represent 3 possible categories, a flaw-free portion, a flaw that does not exceed the depth threshold of 0.1mm, and a flaw that meets or exceeds the depth threshold of 0.1mm. The requirement is to have a binary classifier that classifies each input point as flaw-free or flawed, therefore, the first two categories are labeled as flaw-free '0', and the third type is labeled as flaw '1'. The used UT scans were parsed to create a 75/25 split between flaw-free and flaw inputs. All ultrasonic waves were pre-processed by truncating each to a 100px length down from an average of 1,000px to include only the amplitude peak portion and the data which surrounds it. Fig. 6, Fig. 7, and Fig. 8 illustrate examples of a flaw-free pipe, a flaw that did not meet the depth requirement which is also considered as flaw-free, and a flaw that met or exceeded the depth requirement, respectively. For this prototype, a separation between cross-validation and test datasets has been applied. For cross-validation, 10,000 inputs points were allocated as part of the 60,000 input points training dataset. However, for testing, 18 full UT scan files were used which were not included in the creation of this dataset. These UT test files contained a total of 151,090 inspection points at a resolution of 1mm/2°.

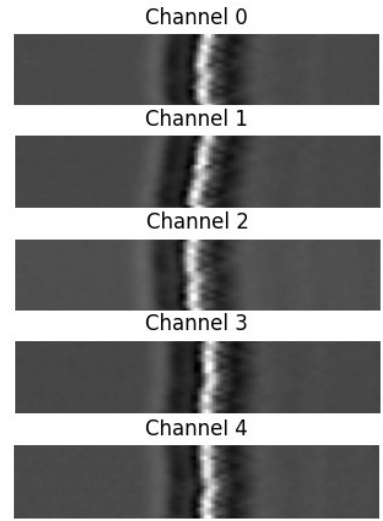

Fig. 6. A flaw-free example (Class '0')

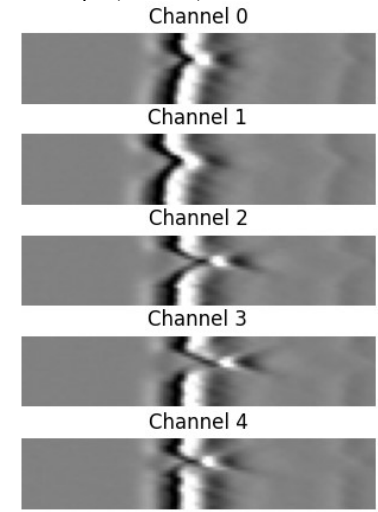

Fig. 7. A flaw that did not meet the depth requirements. (Class '0')

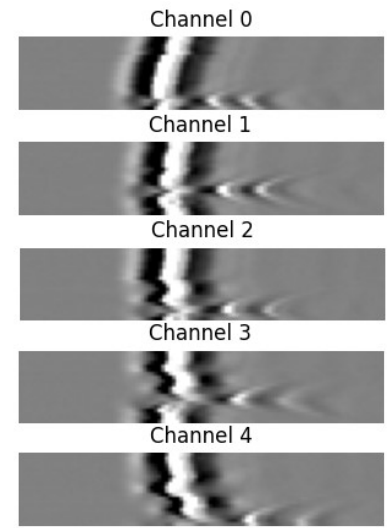

Fig. 8. A flaw that met the depth requirements. (Class '1')

## IV. CNN Model

To develop the required deep learning model, the popular python library "Keras" [22] was used. Using Keras, several CNN architectures were evaluated for the purpose of achieving a binary classification for the existence of flaws based on a resolution of 1mm/2°. Table I illustrates the details of the selected CNN architecture.

TABLE I
TRAINED CNN NETWORK ARCHITECTURE

| ID | Layer | Output Shape | Parameters Count |
|----|-------|--------------|------------------|
| 1 | Input | (100,20,5) | N/A |
| 2 | Conv-2D 300-(5x5) filters with 2x2 stride | (48,8,300) | 37,800 |
| 3 | Conv-2D 300-(5x5) filters with 2x2 stride | (22,8,300) | 2,250,300 |
| 4 | Dense-512 | (512) | 6,758,912 |
| 5 | Dense-128 | (128) | 65,664 |
| 6 | Dense-128 | (128) | 16,512 |
| 7 | Dense-64 | (64) | 8,256 |
| 8 | Dense-10 | (10) | 650 |
| 9 | Output-Dense-2 (Softmax) | (2) | 22 |

As can be seen in the table, the proposed model has two 2D-convolution layers with 300 filters each. Filter sizes of 5x5 were used in both layers with a 2x2 stride. Following the convolution layers, six fully connected layers were used with the sizes of 512, 128, 128, 64, 10, and 2, respectively. The Rectified Linear Units (ReLU) [23] activation function was used in all layers except the output layer where Softmax was used. Also, to avoid overfitting, L2 Regularization [24] and Dropout [25] were used. Table I also details the output shape and parameter count for each layer. Fig. 9 illustrates the proposed CNN architecture.

To appropriately evaluate the performance of the model, confusion matrices are often used [26]. A confusion matrix is an $n \times n$ matrix, where $n$ is the number of output classes for the model. The matrix illustrates the statistics of the actual total vs. predicted total of each output class. In the case of binary classification *(n=2)*, a *2 × 2* confusion matrix can be constructed. This matrix can be used to illustrate True Positives

(TP), True Negatives (TN), False Positives (FP), and False Negatives (FN). Using these metrics, the model's accuracy, sensitivity, and specificity can be calculated using equations (1-3) below.

$$Accuracy = \frac{TP + TN}{TP + TN + FP + FN} \tag{1}$$

$$Sensitivity = \frac{TP}{TP + FN} \tag{2}$$

$$Specificity = \frac{TN}{TN + FP} \tag{3}$$

By referring to the confusion matrix in Table II and based on equations (1-3), the proposed CNN model achieved an accuracy of 93.15%, a sensitivity of 91.17% and a specificity of 93.72% based on the 10,000 cross-validation samples.

TABLE II
CONFUSION MATRIX

| Predicted Actual | No-Flaw (0) | Flaw (1) | Total |
|------------------|-------------|----------|-------|
| No-Flaw (0) | TN: 7281 | FP: 488 | TN + FP: 7769 |
| Flaw (1) | FN: 197 | TP: 2034 | FN + TP: 2231 |
| Total | TN+FN: 7478 | FP+TP: 2522 | TN+FN+FP+TP: 10000 |

To reduce FPs and increase the inspection accuracy, a post-processor was implemented. The post-processor calculates the average depth per channel in each identified flaw point, then the absolute difference between this average and the depth of each waveform is calculated to determine if the identified point meets the depth threshold. The post-processor used an estimated depth measurement of 0.09mm as a threshold instead of the defined 0.1mm. The existence of the chatter imposes a challenge on performing accurate depth measurement. Therefore, using a threshold of 0.09 allows for a more conservative approach and avoids causing FNs especially for flaws that are exactly measured with a depth of 0.1mm by analysts.

Identifying the location of a portion of each flaw is sufficient as manual analysis will be performed to identify the extent and the type of the flaw. Flaws are larger than the inspection resolution of 1mm/2°. This leads to having multiple inspection points per flaw and having one detection point is sufficient to satisfy the requirement. Fig. 10 summarizes the steps proposed in this PoC to automate the detection of flaws in UT scans in production.

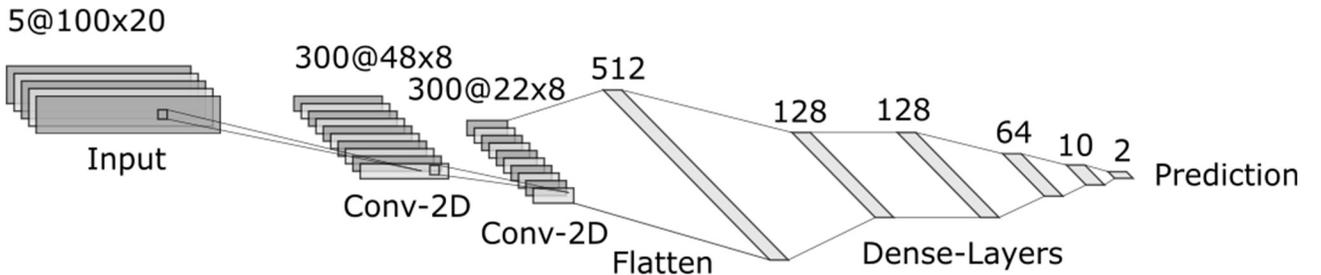

Fig. 9. The proposed CNN model architecture

As presented in Fig. 10, several steps are required to automate the inspection of UT scans. First, all ultrasonic waves which represent a resolution of 0.2mm/0.1° are truncated to 100px in length down from an average of 1000px, then bundles of 100 waves are group together to construct a 5-channel input point with a lower resolution of 1mm/2°. These constructed inspection points are passed to the trained CNN model for prediction. After obtaining the CNN predictions, a post-processor is used to minimize FPs by performing an estimated depth measurement. Finally, an expert analyst can review the identified positive points and determine the extent of each flaw in terms of length and width. The next section will present the test results for different types of UT scans.

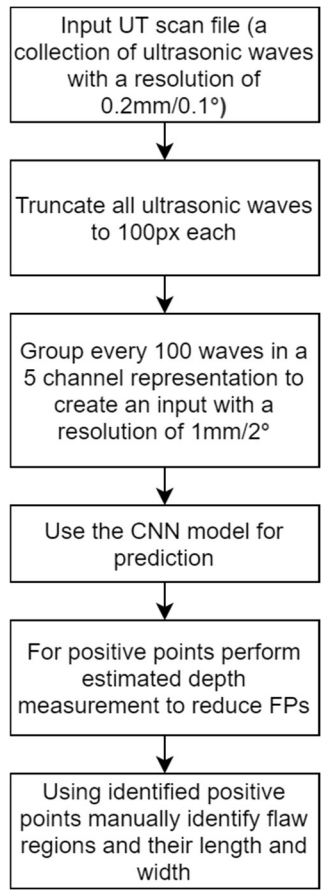

Fig. 10. Summary of the steps used to automate the detection of flaws in fuel channel UT scans

## V. TEST RESULTS

A total of 18 UT full scan files were used to test the proposed model. These scans include different flaw types from various reactors, fuel channels, and outages. These UT scan files contained 151,090 inspection points at a resolution of 1mm/2°. These scans are highly unbalanced as most inspection points are flaw-free. Unlike the cross-validation data, which is part of the training dataset, these test scan files were not parsed to create a balance between flaw-free and flaw points. Testing used UT scans as-is to evaluate how this prototype would work in production where the entire scan file is analyzed. Fig. 11 depicts the spatial representation of 5 test cases by following the process which is illustrated in Fig. 10.

In Fig. 11, the blue dots represent flaw prediction points with a resolution of 1mm/2° predicting a region with a depth of 0.1mm or greater. On the other hand, the red boxes depict the borders of flawed regions that were identified manually by a human analyst. These borders are identified by including all surrounding regions with a depth that is measured as 0.01mm or greater. Therefore, not every single point in the red labeled flawed region is expected to be identified as the CNN is only locating depths of 0.1mm or greater. Flaw regions that don't contain at least one point with a depth of 0.1mm or more are not marked by analysts.

As mentioned earlier, the requirement for this prototype was to detect at least a portion of each flaw with a depth of 0.1mm or greater. This was met in all 18 full scan test cases. Fig. 11 demonstrates the inspection results of 5 full scan test cases representing different flaw types. Fig. 11 (a) and (b) present examples of debris fret flaws. Fig. 11 (c) and (d) present examples of crevice corrosion flaws, while Fig. 11 (e) presents a case of fuel bundle bearing pad fretting (FBBPF) flaw. As can be seen in Fig. 11, in all test cases, one or more positive points were identified in each flaw. Most of the FPs in the examples are borderline flaws that have depths which are slightly below the threshold of 0.1mm.

## VI. CONCLUSION

This paper presented a PoC for automating the detection of flaws in CANDU™ nuclear fuel channel UT scans using deep learning. The PoC aims to reduce nuclear power plant outage operational costs by automating the current manual inspection process for pressure tubes which are an essential part of the reactor's fuel channels. The requirement for this prototype was to identify the location of at least a portion of each flaw region in each scan while minimizing false positives. Using the proposed methodology, 100 ultrasonic waves are grouped together after truncation to create a 5-channel input. The inputs which represent a resolution of 1mm/2° are then passed to a proposed CNN network which performs a binary classification to determine the location of flaw points. Following the CNN, a post-processor is used to reduce FPs by performing an estimated depth measurement for each channel in the identified positive points. By testing the proposed solution using 18 full UT scan files, at least one point in each flaw was identified automatically which allows analysts to quickly identify the location of flawed regions where manual analysis is applied to determine the length and width of each region. Automating the identification of flaws lengths and widths as part of a future research is needed to provide a fully automated solution. The proposed ideas in this manuscript in terms of using lower resolution UT data with CNNs to automate the detection of flaws or fractions can be applied to other pipeline inspections [27-28] and in structural inspection [29-30].

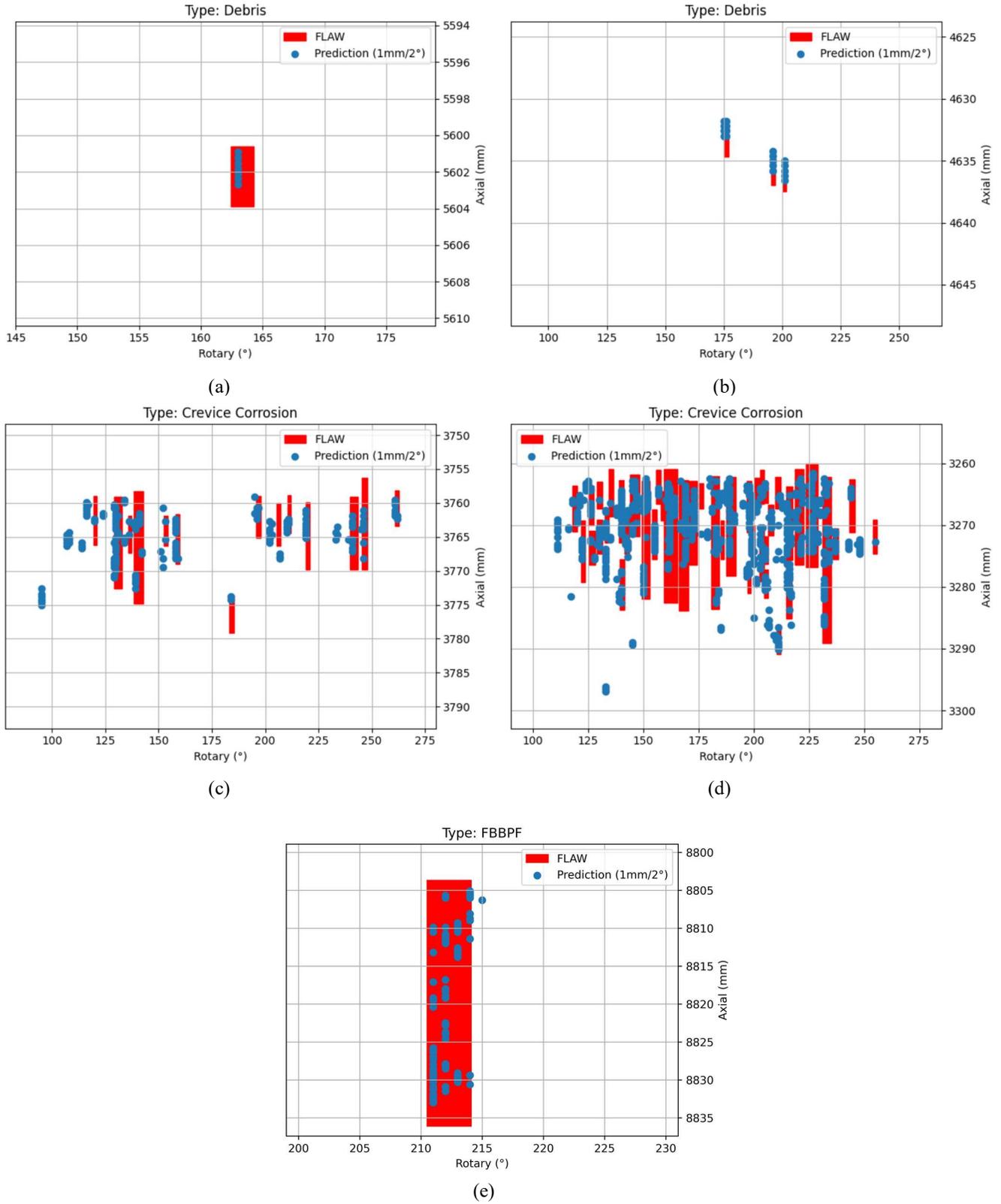

Fig. 11. Examples for test cases.
   (a)   An example for a debris flaw with 1 region
   (b)   An example for a debris flaw with 3 regions
   (c)   An example for a crevice corrosion flaw with 14 regions
   (d)   An example for a crevice corrosion flaw with 44 regions
   (e)   An example for a large FBBPF flaw

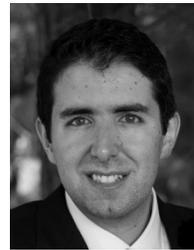

**ISSAM HAMMAD** received the B.Sc degree in computer engineering from Princess Sumaya University for Technology, Amman, Jordan in 2008, the M.A.Sc and the Ph.D. degrees in electrical and computer engineering from Dalhousie University, NS, Canada in 2010 and 2021, respectively. He is currently a Senior Data Scientist in the engineering department at Alithya Digital Technology Corporation, Toronto, ON, Canada. He has more than twelve years of engineering industry experience in machine learning/deep learning, big data, engineering automation, control systems, predictive maintenance, video compression and analysis, cryptography and cybersecurity, cloud computing, distributed computing, embedded systems, custom hardware (FPGA and ASIC), IoT/ IIoT, nuclear software systems, nuclear software quality and standards, full-stack web development, and database design. His current research interests include AI for Inspection and Industrial Automation, Edge AI, Approximate and Stochastic Computing, Hardware–Software co-optimization, and AI for Video/Audio/Ultrasonic Analysis.

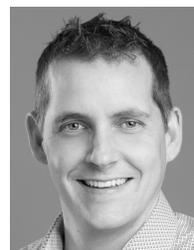

**RYAN SIMPSON** received the B.Sc degree in electrical engineering from Queen's University, Kingston, Ontario in 2005. He currently holds a Professional Engineering license from Professional Engineers Ontario, and a Functional Safety Engineering license from TÜV Rheinland. He has more than 15 years of experience in the Canadian nuclear industry, with specific expertise in the areas of computer system engineering, control systems engineering, and fuel channel inspection and maintenance systems, including robotics and ultrasonic non-destructive evaluation inspections. He is currently working as the VP Consulting Services for the engineering department at Alithya Digital Technology Corporation in Toronto, ON.

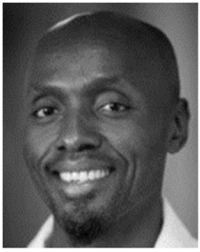

**HIPPOLYTE DJONON TSAGUE** received a B.Sc. and a M.Sc. in Electrical and Electronics Engineering from the University of the Witwatersrand in Johannesburg South Africa in 2003 and 2006 respectively and a Ph.D. in computational Intelligence from the University of Johannesburg South Africa in 2018. His current research interests include machine learning, deep learning, formal methods, signal processing and side channel analysis. He has more than 15 years of industrial experience in embedded development, software development, machine learning and natural language processing. He is currently working as a Senior Data Scientist for the energy department at Alithya Digital Technology Corporation in Toronto, ON.

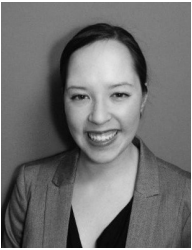

**SARAH HALL** received a B.Eng degree in chemical engineering from McMaster University, Hamilton, Canada in 2016. She has 5 years of experience with projects involving nuclear software quality assurance, machine learning and plant modernization. Sarah is a certified Project Management Professional (PMP), and she is currently working as a project manager for the energy department at Alithya Digital Technology Corporation in Toronto, ON.